\documentclass[10pt,twocolumn,letterpaper]{article}

\usepackage{cvpr}              

\usepackage{tikz}
\usetikzlibrary{arrows.meta,positioning,fit,backgrounds,calc}
\usepackage[accsupp]{axessibility}  

\definecolor{cvprblue}{rgb}{0.21,0.49,0.74}
\usepackage[pagebackref,breaklinks,colorlinks,allcolors=cvprblue]{hyperref}

\def\paperID{00022} 
\def\confName{CVPR}
\def\confYear{2026}

\title{Latent Space Reinforcement Learning for Inverse Material Estimation in Food Fracture Simulation}

\author{Adrian Ramlal \quad
Yuhao Chen \quad
John S. Zelek\\
University of Waterloo\\
{\tt\small \{adrian.ramlal, yuhao.chen1, jzelek\}@uwaterloo.ca}
}

\begin{document}

\twocolumn[{
\maketitle
\begin{center}
\captionsetup{type=figure}
\includegraphics[width=1.0\textwidth]{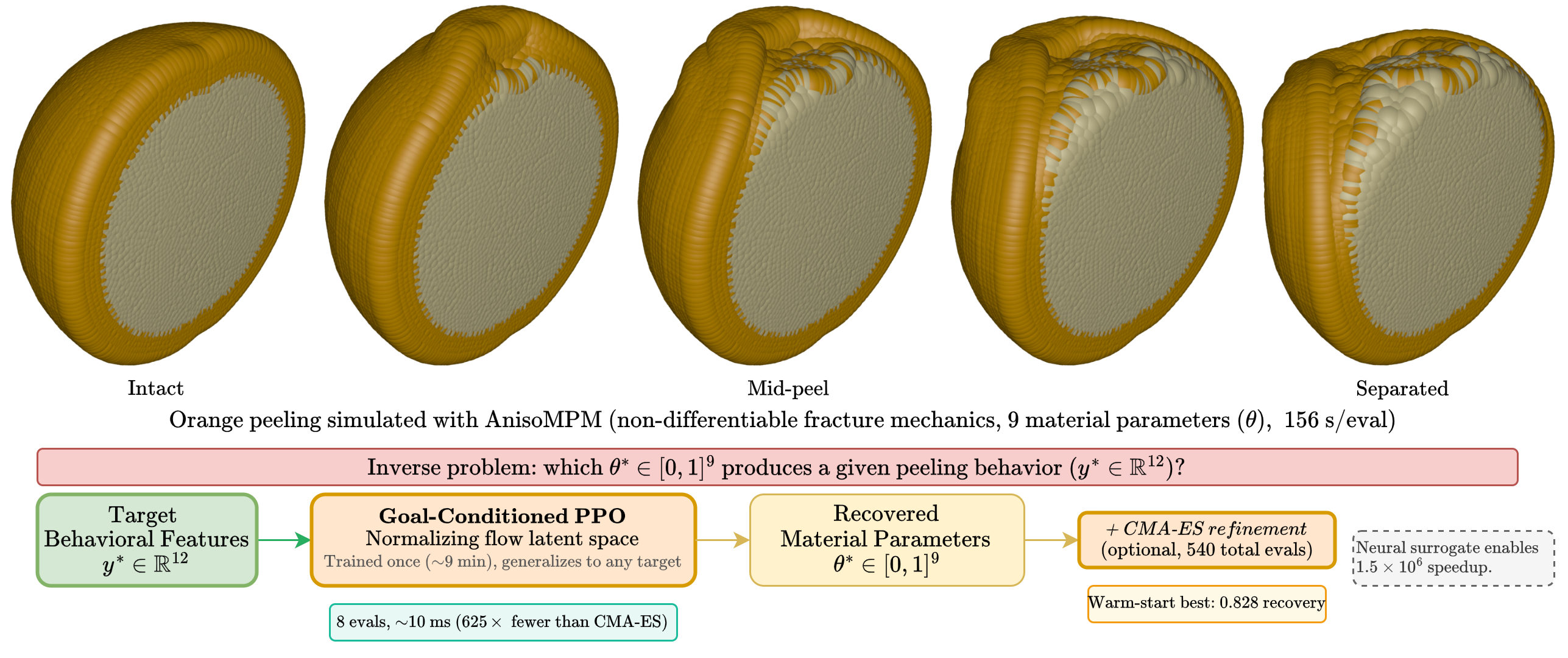}
\captionof{figure}{Orange peeling simulated with AnisoMPM~\cite{wolper2020anisompm}, a non-differentiable fracture mechanics simulator with $9$ material parameters and ${\sim}156$\,s per evaluation. Different parameter settings produce qualitatively different peeling behaviors. Our inverse estimation pipeline. Given target behavioral features $\mathbf{y}^*$ describing a desired peeling outcome, a goal-conditioned PPO policy operating in a normalizing flow latent space recovers the material parameters $\boldsymbol{\theta}^*$ in $8$ surrogate evaluations (${\sim}10$\,ms), $625\times$ fewer than CMA-ES. The policy is trained once (${\sim}9$\,min) and generalizes to arbitrary targets without retraining. A neural surrogate provides a $1.5{\times}10^6$ speedup over the simulator, and optional CMA-ES warm-start refinement ($540$ total evaluations) pushes recovery to $0.828$.}
\label{fig:teaser}
\end{center}
}]

\begin{abstract}
Realistic visual simulation of food manipulation requires accurate material parameters, yet these are difficult to measure directly and vary across the heterogeneous regions of a single food item.
We address the inverse problem of estimating material parameters from a target description of fracture behavior in a non-differentiable continuum damage mechanics simulator.
Using orange peeling as a test case, we train a neural surrogate on 2,000 forward simulations and compare Covariance Matrix Adaptation Evolution Strategy (CMA-ES, a gradient-free evolutionary optimizer) with Proximal Policy Optimization (PPO, a reinforcement learning algorithm) across the original 9-dimensional parameter space and two learned 4-dimensional latent representations.
Since different oranges have different material properties, a practical inverse system must handle arbitrary targets without retraining. We train a goal-conditioned PPO policy that learns a general inverse mapping: given any target description of peeling behavior, the policy produces a material parameter estimate in a single forward pass (8 surrogate evaluations, approximately 10\,ms).
Operating in a normalizing flow latent space with a shared surrogate evaluator, the goal-conditioned policy achieves 0.642 actual recovery when validated through the simulator, outperforming the original parameter space by 23\%.
A warm-start extension that initializes CMA-ES refinement from the policy's output further improves recovery to 0.828 with 540 evaluations.
These findings provide a practical framework for inverse food physics and lay groundwork for vision-driven material identification from video observations of food manipulation.

\end{abstract}    
\section{Introduction}
\label{sec:intro}


Recent advances in visual tracking and 3D reconstruction have made it increasingly practical to observe and quantify the physical behavior of food during manipulation, from robotic cutting~\cite{heiden2021disect, xu2023roboninja, beltran2024sliceit} to deformable object handling~\cite{shi2023robocook} and physics-aware visual synthesis~\cite{wu2025fruitninja, xie2024physgaussian}. A natural next step is to close the loop: given visual observations of food behavior, infer the underlying physical material parameters that would reproduce that behavior in simulation. Such \emph{inverse material estimation} would enable building digital twins of food items from video, calibrating simulators for robotic manipulation planning, and generating physically plausible training data for food recognition models.

This paper investigates the inverse estimation component of this pipeline. Assuming that behavioral features (such as separation distances, damage patterns, and fracture timing) have been extracted from observations, we ask: how can we efficiently recover the material parameters that produce matching behavior in a fracture simulator? We focus on this inverse step because, while visual tracking and feature extraction have mature solutions~\cite{wu2025fruitninja}, the inverse estimation problem remains open for non-differentiable fracture simulators where gradient-based inversion is impossible.


We study this problem using orange peeling simulated with AnisoMPM~\cite{wolper2020anisompm}, a Material Point Method (MPM) simulator with continuum damage mechanics and anisotropic fiber reinforcement (Figure~\ref{fig:teaser}). The simulator has a $9$-dimensional material parameter space, is non-differentiable due to topological discontinuities during crack formation, takes minutes per evaluation, and exhibits chaotic sensitivity to parameters. Although recent work has demonstrated inverse estimation for differentiable simulators~\cite{ma2023nclaw, heiden2021disect} or from direct force measurements~\cite{beltran2024sliceit}, these approaches cannot be applied when the simulator itself is non-differentiable.

Our approach begins with a neural surrogate trained on $2{,}000$ forward simulations, providing a $10^6\times$ speedup that enables iterative optimization. We establish that Covariance Matrix Adaptation Evolution Strategy (CMA-ES)~\cite{hansen2001cmaes}, a standard gradient-free optimizer, can recover material parameters but requires thousands of evaluations per target, and that Proximal Policy Optimization (PPO)~\cite{schulman2017ppo}, a reinforcement learning (RL) algorithm, achieves comparable results, validating RL for inverse fracture physics.

The key limitation of both CMA-ES and per-target PPO is that they must be re-run for each new target. Since different oranges have different material properties (e.g., varying skin thickness, fiber density, and flesh firmness depending on variety and ripeness), a practical inverse system must handle arbitrary targets without retraining. We address this by training a \emph{goal-conditioned} PPO policy that conditions on the target features and learns a general inverse mapping. Once trained (${\sim}9$ minutes, one-time), this policy produces parameter estimates for any new target in $8$ evaluations (${\sim}10$\,ms). We evaluate the policy across three parameter representations: the original $9$-dimensional space and two $4$-dimensional latent spaces (VAE~\cite{kingma2014vae}, normalizing flow~\cite{dinh2017realnvp}). VAEs and normalizing flows are the two dominant generative models used for latent space optimization~\cite{gomez2018automatic, tripp2020sample, lee2025nfbo, maus2022lolbo}; we include both to evaluate whether the bijective property of flows provides advantages over the stochastic encoding of VAEs, as suggested by prior work~\cite{lee2025nfbo}. When all methods use the same surrogate evaluator, the flow latent space achieves the best recovery ($0.642$ actual), outperforming the original $9$D space by $23\%$.


Our contributions are:
\begin{itemize}
    \item A goal-conditioned RL policy that produces inverse material estimates for arbitrary targets in $8$ evaluations (${\sim}10$\,ms), achieving $0.642$ actual recovery without per-target retraining.
    \item A systematic comparison of evolutionary search and RL across the original parameter space and learned latent spaces for non-differentiable food fracture simulation.
    \item An ablation showing that evaluation design matters: latent methods using their own prediction heads perform worse than those using a shared, higher-accuracy surrogate, due to surrogate exploitation.
    \item A warm-start hybrid that initializes CMA-ES from the policy's output, achieving the best overall recovery ($0.828$, $540$ evaluations).
\end{itemize}

\section{Related Work}
\label{sec:related}

\paragraph{Food physics simulation and manipulation.}
Physics-based simulation of food materials spans robotics and computer graphics. Ding~et~al.~\cite{ding2019thermomechanical} introduced a thermomechanical MPM for simulating baking. DiSECt~\cite{heiden2021disect} proposed a differentiable simulator for cutting and used Bayesian inference to calibrate material parameters from force measurements. RoboNinja~\cite{xu2023roboninja} trained adaptive cutting policies with differentiable MPM, and RoboCook~\cite{shi2023robocook} addressed long-horizon elasto-plastic manipulation. CulinaryCut-VLAP~\cite{koh2026culinarycut} coupled vision-language models with MPM for food cutting. Most closely related to our inverse estimation objective, SliceIt!~\cite{beltran2024sliceit} calibrated food material parameters from real-world force measurements in a dual-simulator pipeline for robotic slicing. On the visual side, FruitNinja~\cite{wu2025fruitninja} generated realistic interior textures for 3D Gaussian Splatting fruit models, and PhysGaussian~\cite{xie2024physgaussian} integrated MPM with Gaussian Splatting for physics-driven dynamics. These works either assume differentiable simulators, calibrate from direct force measurements, or specify parameters manually. We address the inverse problem of discovering material parameters for a \emph{non-differentiable} fracture simulator from behavioral feature targets alone.

\paragraph{Material Point Method, fracture, and inverse problems.}
The Material Point Method combines Lagrangian particles with an Eulerian grid to handle large deformations and fracture. Wolper~et~al.~\cite{wolper2019cdmpm} introduced continuum damage mechanics into MPM, and AnisoMPM~\cite{wolper2020anisompm} extended this to anisotropic damage with directional fiber reinforcement. These simulators are non-differentiable: topological discontinuities during fracture preclude gradient computation through the simulation. This distinguishes our setting from differentiable physics frameworks such as DiffTaichi~\cite{hu2020difftaichi}, FluidLab~\cite{xian2023fluidlab}, and PlasticineLab~\cite{huang2021plasticinelab}, where gradients enable direct inverse solving. NCLaw~\cite{ma2023nclaw} demonstrated that neural constitutive laws can be learned within a differentiable MPM, but this requires backpropagation through the simulator, which is fundamentally unavailable for fracture mechanics.

\paragraph{Latent space optimization and reinforcement learning.}
Optimization in learned latent spaces was pioneered by G\'{o}mez-Bombarelli~et~al.~\cite{gomez2018automatic} for molecular design, followed by weighted retraining~\cite{tripp2020sample} to focus generative models on high-scoring regions. LOL-BO~\cite{maus2022lolbo} added trust regions to latent Bayesian optimization, and Lee~et~al.~\cite{lee2025nfbo} identified a value discrepancy in VAE-based optimization that normalizing flows eliminate through bijective mappings. In robotics, PULSE~\cite{luo2024pulse} demonstrated that distilling motor skills into a latent space enables efficient hierarchical RL, and OmniGrasp~\cite{luo2024omnigrasp} applied this for dexterous grasping. We adapt the latent-space-for-control principle from the motor domain to parameter identification.

\paragraph{Surrogate-based optimization and amortized inference.}
Training a neural surrogate on a pre-computed dataset and optimizing against it is formalized as offline model-based optimization (MBO), benchmarked through Design-Bench~\cite{trabucco2022designbench}. Surrogate exploitation, where the optimizer discovers inputs scoring highly on the surrogate but poorly under the true objective, is a well-known failure mode addressed by conservative training~\cite{trabucco2021coms} and robustness constraints~\cite{yu2021roma}. CMA-ES~\cite{hansen2001cmaes} is the standard gradient-free optimizer in this setting~\cite{fischer2024zerograds}. Amortized inference trains a single model to solve a distribution of inverse problems~\cite{cranmer2020frontier, amos2023tutorial}. Ardizzone~et~al.~\cite{ardizzone2019inn} used invertible neural networks for physics inverse problems. Goal-conditioned policies~\cite{schaul2015universal} generalize across objectives by conditioning on a target specification.

\section{Method}
\label{sec:method}
\begin{figure*}
    \centering
    \includegraphics[width=1.0\linewidth]{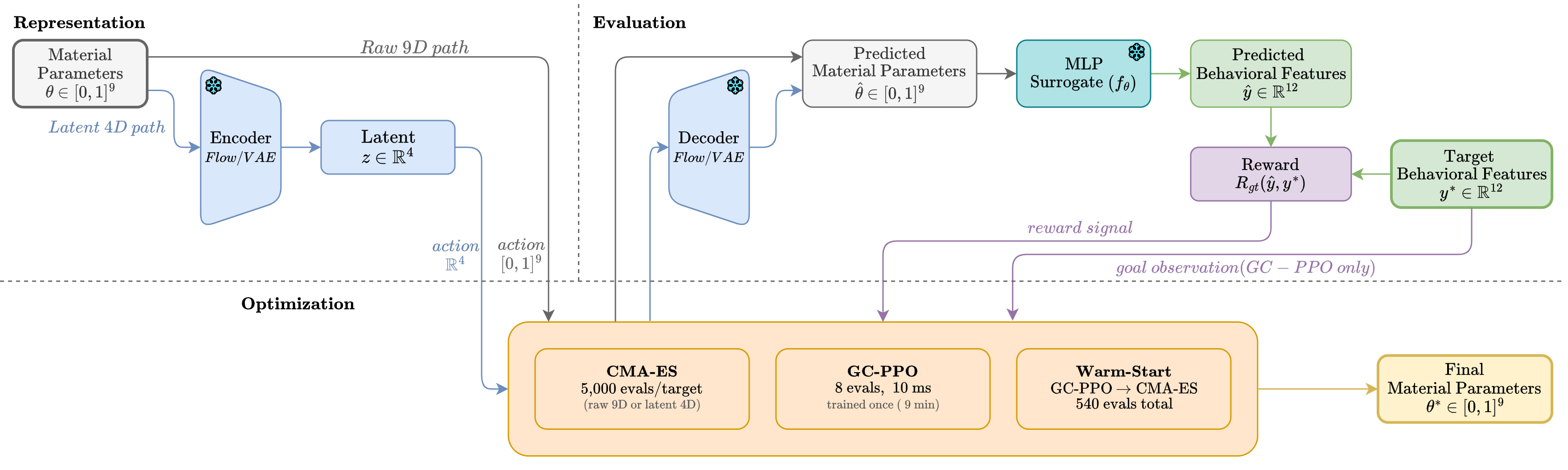}
    \caption{Method overview. At inference, all pretrained components (encoder, decoder, surrogate) are frozen. \textbf{Representation} (top left): parameters are optimized directly or compressed to a $4$D latent code. \textbf{Optimization} (bottom): CMA-ES searches per-target; GC-PPO conditions on target features $\mathbf{y}^*$ as a goal observation and generalizes across targets; warm-start combines both. \textbf{Evaluation} (top right): proposals are decoded to parameters, passed through the frozen MLP surrogate ($R^2{=}0.937$), and scored against the target. The decode-then-evaluate design ensures all methods use the same evaluator, isolating representation quality from surrogate quality.}
    \label{fig:pipeline}
\end{figure*}

We present a pipeline for inverse food fracture parameter estimation from a non-differentiable simulator. The pipeline consists of four components: simulation and dataset generation (Sec.~\ref{sec:simulation}), behavioral feature extraction and surrogate modeling (Sec.~\ref{sec:surrogate}), latent space models for parameter compression (Sec.~\ref{sec:latent}), and optimization methods including a goal-conditioned policy and warm-start hybrid (Sec.~\ref{sec:gc}--\ref{sec:warmstart}). Figure~\ref{fig:pipeline} provides an overview.

\subsection{Simulation and Parameter Space}
\label{sec:simulation}

\begin{figure}[t]
\centering
\includegraphics[width=1.0\linewidth]{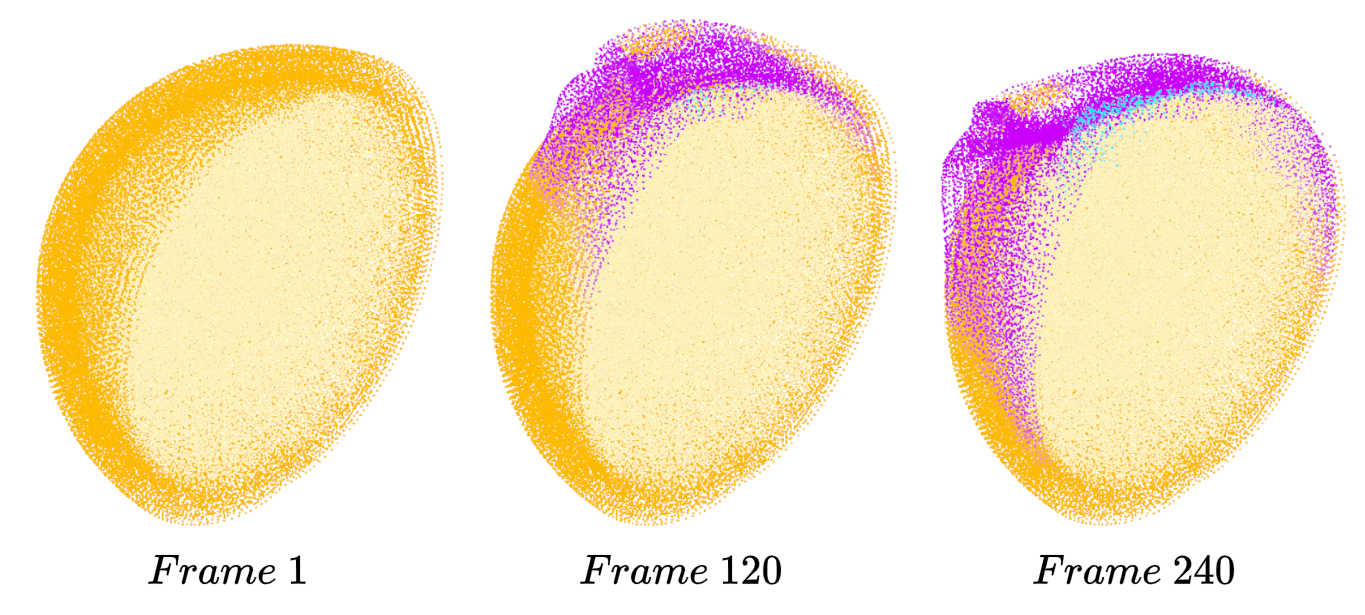}
\caption{AnisoMPM simulation of orange peeling at three stages. Particles are colored by region and damage: orange (intact skin), yellow (intact flesh), purple (damaged skin), cyan (damaged flesh). The anisotropic fiber-reinforced skin resists tearing along preferred orientations, producing heterogeneous fracture patterns that depend sensitively on the $9$ material parameters (Table~\ref{tab:params}).}
\label{fig:sim_scene}
\end{figure}

We use AnisoMPM~\cite{wolper2020anisompm} to model orange peeling. AnisoMPM extends MPM~\cite{wolper2019cdmpm} with anisotropic damage mechanics and directional fiber reinforcement, necessary because orange skin exhibits directional fiber structure that resists tearing preferentially along certain orientations. Standard isotropic MPM cannot capture this behavior. The simulation represents an orange half with distinct skin and interior (flesh) material regions, where a gripper pulls the skin away from the interior along a fixed arc trajectory (Figure~\ref{fig:sim_scene}). The skin is modeled with anisotropic fiber reinforcement controlled by the fiber stiffness scale $f_s$. The interior uses isotropic damage mechanics with no fiber reinforcement, representing the softer flesh. Each simulation produces $240$ frames of ${\sim}50{,}000$ particles and runs for ${\sim}156$\,s on $8$ CPU threads.

The simulation is controlled by 9 physical parameters (Table~\ref{tab:params}). Two parameters are shared across regions: Young's modulus $E$ and density $\rho$. Each region has independent fracture threshold $\sigma$, damage regularization $\eta$, and residual stress $r$. The skin additionally has a fiber stiffness scale $f_s$ controlling anisotropic reinforcement along material fibers. All parameters are normalized to $[0,1]$ for optimization, with log-scale transformation for the six that span multiple orders of magnitude.

\begin{table}[t]
\centering
\caption{The 9 material parameters, with physical ranges and region assignments. Six parameters (marked $\dagger$) use log-scale normalization.}
\label{tab:params}
\small
\begin{tabular}{lccc}
\toprule
\textbf{Parameter} & \textbf{Symbol} & \textbf{Range} & \textbf{Region} \\
\midrule
Young's modulus$^\dagger$ (Pa) & $E$ & $[50, 8000]$ & shared \\
Density               & $\rho$ & $[0.5, 10]$ & shared \\
Fracture threshold    & $\sigma_{\text{s}}$ & $[0.05, 0.5]$ & skin \\
Damage regularization$^\dagger$ & $\eta_{\text{s}}$ & $[0.01, 1.0]$ & skin \\
Residual stress$^\dagger$       & $r_{\text{s}}$ & $[0.001, 0.1]$ & skin \\
Fiber stiffness$^\dagger$       & $f_s$ & $[1, 100]$ & skin \\
Fracture threshold    & $\sigma_{\text{i}}$ & $[0.1, 0.9]$ & interior \\
Damage regularization$^\dagger$ & $\eta_{\text{i}}$ & $[0.01, 1.0]$ & interior \\
Residual stress$^\dagger$       & $r_{\text{i}}$ & $[0.001, 0.1]$ & interior \\
\bottomrule
\end{tabular}
\end{table}

We generate $N = 2{,}000$ simulations by sampling material parameters uniformly across the normalized parameter space, totaling $87$ hours of compute. The simulator is deterministic (verified with zero variance across $5$ identical runs), so any discrepancy between surrogate-predicted and simulator-validated rewards is attributable entirely to surrogate inaccuracy.

\subsection{Feature Extraction and Neural Surrogate}
\label{sec:surrogate}
Running the simulator is too expensive for direct use in optimization ($2.6$ minutes per evaluation). We address this by defining a compact set of behavioral features that summarize each simulation's outcome, then training a fast neural surrogate that predicts these features from material parameters.


We design $12$ aggregate behavioral features that capture macroscopic peeling characteristics. These features are hand-crafted from domain knowledge of the peeling process and computed from the raw particle data by reading every 10th frame (yielding ${\sim}24$ frames from $240$) and splitting particles into skin and interior groups using a per-particle label stored in the simulation output. The features capture four aspects of peeling behavior: separation dynamics (final centroid distance between skin and interior, peak separation rate, fracture onset frame, separation half-time), skin behavior (spatial dispersion, detachment fraction), interior behavior (mean and maximum displacement, spatial dispersion), and material damage (mean damage scalar in skin and interior regions).

\begin{figure*}
\centering
\includegraphics[width=1.0\linewidth]{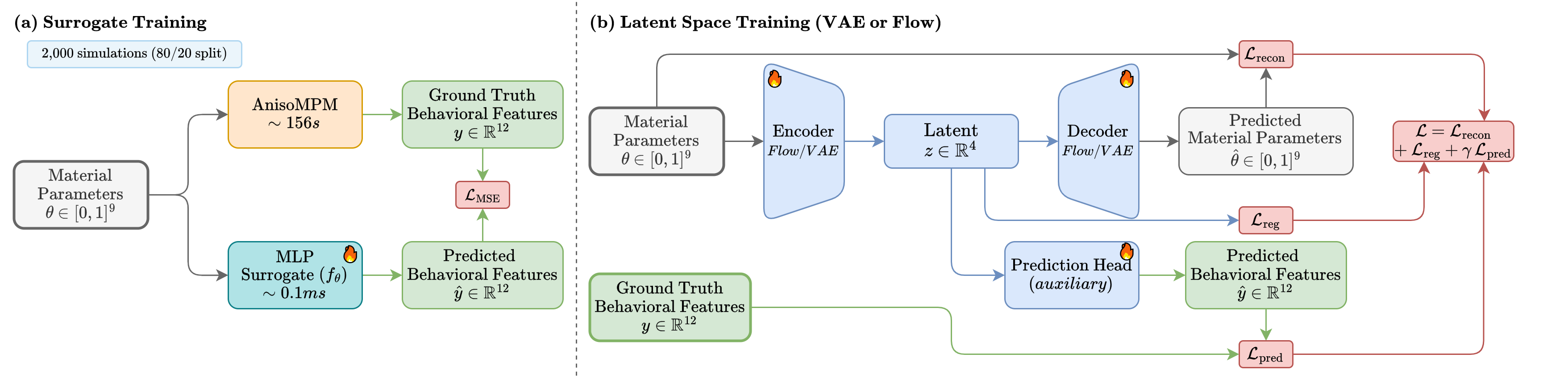}
\caption{Offline training (trainable modules indicated in figure). \textbf{(a)}~The MLP surrogate learns to predict behavioral features from material parameters, providing a $1.5{\times}10^6$ speedup over the simulator. \textbf{(b)}~The latent model (VAE or flow) compresses the $9$D parameter space to $4$D via three jointly trained objectives: parameter reconstruction ($\mathcal{L}_{\text{recon}}$), latent regularization ($\mathcal{L}_{\text{reg}}$), and behavioral feature prediction ($\mathcal{L}_{\text{pred}}$) via an auxiliary head. The feature prediction term organizes the latent space so that nearby codes produce similar peeling behaviors, enabling efficient RL navigation.}
\label{fig:training}
\end{figure*}

We train an MLP surrogate $f_\theta: [0,1]^9 \to \mathbb{R}^{12}$ predicting features from parameters (Figure~\ref{fig:training}a). The architecture is a multi-layer perceptron with hidden dimensions ($256$, $256$, $128$), a skip connection concatenating the outputs of the first two layers into a $512$-dimensional input to the third, batch normalization, ReLU activations, and dropout ($0.1$). Training uses MSE loss on standardized features with an 80/20 split ($1{,}600$ / $400$ samples), Adam optimizer (learning rate $10^{-3}$, weight decay $10^{-4}$), and early stopping with patience $50$. The surrogate achieves $R^2 = 0.937$ overall (per-feature range: $0.854$ to $0.991$) and evaluates in ${\sim}0.1$\,ms, a $1.5 \times 10^6$ speedup over the simulator.

\paragraph{Reward function.}
We define a recovery reward measuring how well predicted features match a target:
\begin{equation}
    R_{\text{gt}}(\mathbf{y}, \mathbf{y}^*) = \exp\!\Biggl(-\kappa\sqrt{\frac{1}{M}\sum_{k=1}^{M} \biggl(\frac{y_k - y^*_k}{s_k}\biggr)^{\!2}}\;\Biggr),
\label{eq:gt_reward}
\end{equation}
where $\mathbf{y}^*$ is the target feature vector, $s_k = p_{95,k} - p_{5,k}$ is the percentile-based scale from the dataset, $M{=}12$ is the number of features, and $\kappa{=}3$ controls the sharpness of the reward around the target. A value of $R_{\text{gt}}{=}1$ indicates perfect feature matching.

\subsection{Latent Space Models}
\label{sec:latent}

We compare two generative models that compress the $9$D parameter space to $4$D, providing a reduced action space for RL (Figure~\ref{fig:training}b). Both models are trained with a multi-task objective combining parameter reconstruction with behavioral feature prediction, so the latent space is shaped not only by parameter geometry but also by peeling behavior.

\paragraph{Variational autoencoder (VAE).}
The encoder maps parameters through hidden layers of $128$ and $64$ units to a mean and log-variance of a $4$D Gaussian, from which latent codes are sampled via the reparameterization trick~\cite{kingma2014vae}. The decoder maps back through $64$ and $128$ units to reconstructed parameters (bounded by sigmoid). The training loss is:
\begin{equation}
    \mathcal{L}_{\text{VAE}} = \mathcal{L}_{\text{recon}} + \beta \, D_{\text{KL}} + \gamma \, \mathcal{L}_{\text{pred}},
\label{eq:vae_loss}
\end{equation}
where $\mathcal{L}_{\text{recon}} = \|\boldsymbol{\theta} - \hat{\boldsymbol{\theta}}\|^2$ is the parameter reconstruction loss, $D_{\text{KL}} = D_{\text{KL}}(q(\mathbf{z}|\boldsymbol{\theta}) \| p(\mathbf{z}))$ regularizes the latent distribution toward a standard Gaussian prior, and $\mathcal{L}_{\text{pred}} = \|\mathbf{y} - \hat{\mathbf{y}}\|^2$ is a feature prediction loss from an auxiliary head that encourages the latent space to cluster parameter configurations producing similar peeling behavior. We set $\beta{=}0.1$ ($50$-epoch linear warmup) and $\gamma{=}1.0$. Reconstruction MSE: $0.007$.

\paragraph{Normalizing flow.}
A RealNVP-based flow~\cite{dinh2017realnvp} with $6$ affine coupling layers (alternating binary masks, hidden dimension $64$, scale networks bounded by tanh) provides a bijective mapping within the latent space. Since the parameter dimension ($9$) exceeds the latent dimension ($4$), learned projections with ReLU activations map $9 \to 64 \to 4$ before the flow and $4 \to 64 \to 9$ after it. The training loss is:
\begin{equation}
    \mathcal{L}_{\text{Flow}} = \mathcal{L}_{\text{recon}} + \alpha \, \mathcal{L}_{\text{flow}} + \gamma \, \mathcal{L}_{\text{pred}},
\label{eq:flow_loss}
\end{equation}
where $\mathcal{L}_{\text{flow}} = \tfrac{1}{2}\|\mathbf{z}\|^2 - \log|\det J|$ is the negative log-likelihood under the flow (with Jacobian $J$), encouraging the latent distribution to match a standard Gaussian. We set $\alpha{=}0.1$ ($50$-epoch warmup) and $\gamma{=}1.0$. Although the flow achieves $8\times$ worse reconstruction than the VAE (MSE $0.059$ vs.\ $0.007$), its bijective mapping avoids the reconstruction-induced objective inconsistencies that affect VAEs~\cite{lee2025nfbo}.

\paragraph{Evaluation pipeline.} When optimizing in a latent space, the policy proposes a latent code $\mathbf{z}$, which is decoded to material parameters $\hat{\boldsymbol{\theta}} = \text{clip}(\text{decode}(\mathbf{z}), 0, 1)$, and the features are predicted by the MLP surrogate: $\hat{\mathbf{y}} = f_\theta(\hat{\boldsymbol{\theta}})$. This decode-then-evaluate pipeline ensures all methods, regardless of representation space, use the same high-accuracy evaluator ($R^2{=}0.937$). Both latent models also produce auxiliary prediction heads ($R^2{=}0.854$ for VAE, $R^2{=}0.913$ for flow) as a byproduct of the multi-task training. An alternative approach, common in latent space optimization~\cite{gomez2018automatic, maus2022lolbo, lee2025nfbo}, uses these heads directly instead of decoding; we evaluate this alternative in Sec.~\ref{sec:ablation}.

\subsection{Goal-Conditioned PPO}
\label{sec:gc}

Per-target methods (CMA-ES, PPO) must restart for each new target. Goal-conditioned RL~\cite{schaul2015universal} addresses this by conditioning a policy on a target specification, enabling generalization across objectives. We apply this to inverse material estimation: a single policy $\pi_\omega$ conditions on the target features $\mathbf{y}^*$ and learns a general inverse mapping. At each training episode, a target is sampled uniformly from the training partition with additive Gaussian noise ($\sigma{=}0.05$ in normalized feature space) for robustness. The observation is:
\begin{equation}
    \mathbf{o}_t = \bigl[\,\mathbf{s}_t,\; \hat{\mathbf{y}}_t,\; \mathbf{y}^*,\; R_{\text{gt}}(\hat{\mathbf{y}}_t, \mathbf{y}^*),\; t/T\,\bigr],
\end{equation}
where $\mathbf{s}_t$ is the current state ($\boldsymbol{\theta} \in [0,1]^9$ or $\mathbf{z} \in [-3,3]^4$). The agent refines $\mathbf{s}_t$ over $T{=}8$ steps via $\mathbf{s}_{t+1} = \text{clip}(\mathbf{s}_t + 0.5 \cdot \mathbf{a}_t)$, with sparse reward: zero during intermediate steps, and the best $R_{\text{gt}}$ at the final step. Once trained (${\sim}9$ minutes, one-time), the policy handles any new target in $8$ surrogate evaluations (${\sim}10$\,ms). All PPO variants use learning rate $3{\times}10^{-4}$, entropy coefficient $0.01$, clip range $0.2$, $\gamma{=}0.99$, $\lambda_{\text{GAE}}{=}0.95$, $256$ steps per update, batch size $64$, $10$ SGD epochs per update, and an MLP policy with two hidden layers of $256$ units. Training runs for $500$K timesteps across $8$ parallel environments.

\subsection{Warm-Start Hybrid}
\label{sec:warmstart}

The goal-conditioned policy provides rapid but approximate estimates, limited by its $8$-step horizon. We refine its output by decoding the policy's proposal to material parameters in $[0,1]^9$ and initializing CMA-ES at that point with a tight step size ($\sigma_0{=}0.1$), population size $20$, and $500$ evaluations. For latent-space policies (GC-PPO Flow, GC-PPO VAE), the latent code is first decoded through the respective model's decoder; for GC-PPO Raw, the output is already in $[0,1]^9$. CMA-ES always operates in the original $9$D space regardless of which policy produced the initialization. The total cost per target is $540$ evaluations ($40$ policy rollout + $500$ CMA-ES). The GC-PPO component requires no retraining for new targets; only the CMA-ES refinement runs per target.

\subsection{Validation Protocol}
\label{sec:validation}

We validate all methods by re-running discovered parameters through the actual AnisoMPM simulator, extracting features from the simulation output, and computing $R_{\text{gt}}$ against the target. The \emph{reality gap} $\Delta = R_{\text{surr}} - R_{\text{actual}}$ quantifies surrogate exploitation: larger gaps indicate the optimizer has found solutions that score well on the surrogate but transfer poorly to the real simulator. We evaluate on $20$ diverse held-out targets selected via farthest-point sampling in feature space from the test partition, and validate a subset of $10$ through the simulator. Since all methods evaluate the same surrogate (${\sim}0.13$\,ms per call), the number of surrogate evaluations directly reflects computational cost.



%

\section{Experiments}
\label{sec:experiments}

\subsection{Is RL Viable for Inverse Fracture Physics?}
\label{sec:rl_viable}

Before developing the goal-conditioned pipeline, we establish that RL can match the standard evolutionary baseline for inverse food fracture. We define two reward functions over the behavioral features that represent different peeling scenarios: \emph{firm peel}, simulating a firm-skinned orange where the objective is clean skin separation with minimal interior disturbance, and \emph{ripe peel}, simulating a ripe orange where the priority is preserving the delicate interior while still achieving adequate separation. These reward functions are weighted sums of percentile-normalized feature terms. CMA-ES is configured with population size $20$, $\sigma_0{=}0.3$ in $[0,1]^9$, and a budget of $5{,}000$ evaluations (sufficient for convergence in this $9$D space). Each per-target PPO agent observes $\mathbf{o}_t = [\mathbf{s}_t,\, \hat{\mathbf{y}}_t,\, R(\hat{\mathbf{y}}_t),\, t/T]$ where $R$ is the fixed reward function (firm or ripe peel), and refines the state over $T{=}8$ steps.

Table~\ref{tab:per_target} reports simulator-validated results. PPO matches or exceeds CMA-ES: PPO Raw achieves $0.712$ on firm peel versus $0.694$ for CMA-ES Raw. CMA-ES VAE collapses to $0.375$ on firm peel, an early indication that unconstrained search in the VAE latent space is vulnerable to surrogate exploitation. This confirms that RL is a viable optimizer for inverse fracture physics, but like CMA-ES, a separate PPO policy must be trained for each new reward function (${\sim}10$ minutes per policy). This limitation motivates the goal-conditioned approach.

\begin{table}[t]
\centering
\caption{Per-target optimization (simulator-validated). Firm and ripe peel represent different peeling scenarios. PPO matches CMA-ES, validating RL for inverse fracture physics. Best in \textbf{bold}, second best \underline{underlined}.}
\label{tab:per_target}
\small
\begin{tabular}{llcc}
\toprule
\textbf{Method} & \textbf{Space} & \textbf{Firm} $\uparrow$ & \textbf{Ripe} $\uparrow$ \\
\midrule
CMA-ES & Raw 9D & $\underline{0.694}$ & $0.725$ \\
CMA-ES & Flow 4D & $0.674$ & $0.725$ \\
CMA-ES & VAE 4D & $0.375$ & $\mathbf{0.731}$ \\
\midrule
PPO & Raw 9D & $\mathbf{0.712}$ & $0.721$ \\
PPO & Flow 4D & $0.674$ & $0.716$ \\
PPO & VAE 4D & $0.682$ & $\underline{0.729}$ \\
\bottomrule
\end{tabular}
\end{table}

\subsection{Goal-Conditioned Inverse Recovery}
\label{sec:gc_exp}

Table~\ref{tab:gc} reports the main result. The top block shows CMA-ES baselines, which optimize independently per target with $5{,}000$ evaluations each. CMA-ES Raw achieves the highest absolute recovery ($0.781$) but at significant per-target cost. CMA-ES in latent spaces suffers from severe surrogate exploitation: CMA-ES VAE achieves only $0.028$ actual recovery despite a surrogate score of $0.498$.

The bottom block shows goal-conditioned policies that generalize across targets through the decode-then-evaluate pipeline (Sec.~\ref{sec:latent}). GC-PPO Flow achieves the best goal-conditioned recovery ($0.642$, $82\%$ of CMA-ES Raw) with $625\times$ fewer evaluations and the smallest gap ($\Delta{=}0.012$) of any method. It outperforms the original $9$D space ($0.523$) by $23\%$, indicating that the flow's learned $4$D manifold provides better structure for RL navigation, consistent with the value discrepancy analysis of Lee~et~al.~\cite{lee2025nfbo}.

All goal-conditioned methods produce estimates in $8$ evaluations (${\sim}10$\,ms) with no per-target retraining. Training takes ${\sim}9$ minutes (one-time) and generalizes to arbitrary targets. By contrast, CMA-ES requires $5{,}000$ evaluations per target, and per-target PPO requires ${\sim}10$ minutes of training for each new reward function. The goal-conditioned policy's training cost is amortized across all future targets, with a break-even point of $3$--$5$ targets compared to per-target CMA-ES.

\begin{table}[t]
\centering
\caption{Inverse recovery on $10$ held-out targets (simulator-validated). Top: CMA-ES baselines (per-target, $5{,}000$ evals each). Bottom: goal-conditioned policies (generalized, $8$ evals each). All GC-PPO variants use the MLP surrogate via decode-then-evaluate.}
\label{tab:gc}
\small
\begin{tabular}{lcccc}
\toprule
\textbf{Method} & \textbf{Evals} & \textbf{Surr.} & \textbf{Actual} $\uparrow$ & $\boldsymbol{\Delta}$ $\downarrow$ \\
\midrule
CMA-ES Raw & $5{,}000$ & $0.922$ & $\mathbf{0.781}$ & $0.141$ \\
CMA-ES Flow & $5{,}000$ & $0.878$ & $0.552$ & $0.327$ \\
CMA-ES VAE & $5{,}000$ & $0.498$ & $0.028$ & $0.470$ \\
\midrule
GC-PPO Raw & $8$ & $0.531$ & $0.523$ & $\mathbf{0.008}$ \\
GC-PPO Flow & $8$ & $0.654$ & $\underline{0.642}$ & $\underline{0.012}$ \\
GC-PPO VAE & $8$ & $0.510$ & $0.485$ & $0.025$ \\
\bottomrule
\end{tabular}
\end{table}

\subsection{Warm-Start Refinement}
\label{sec:warmstart_exp}

The goal-conditioned policy provides instant estimates but is limited by its $8$-step horizon. We refine its output by initializing CMA-ES at the decoded material parameters with $500$ additional evaluations in $[0,1]^9$. Table~\ref{tab:warmstart} reports simulator-validated results alongside two CMA-ES baselines: one with the same $540$-evaluation budget (from random center initialization) and the standard $5{,}000$-evaluation configuration from Table~\ref{tab:gc}. The best warm-start variant (WS VAE, $0.828$) outperforms both CMA-ES baselines. WS Raw ($0.808$) also outperforms both, while WS Flow ($0.773$) underperforms CMA-ES $540$ ($0.788$), showing that the benefit depends on the policy representation. The warm-start gaps ($\Delta{=}0.034$--$0.095$) are smaller than CMA-ES $5{,}000$ ($\Delta{=}0.141$), indicating that policy initialization steers CMA-ES away from surrogate-overestimated regions. All warm-start computations take ${\sim}0.16$\,s on the surrogate, making the per-target cost dominated by the single simulator validation (${\sim}156$\,s).

\begin{table}[t]
\centering
\caption{Warm-start and CMA-ES comparison ($10$ held-out targets, simulator-validated). WS variants use the respective GC-PPO policy for initialization, then CMA-ES refines in $[0,1]^9$.}
\label{tab:warmstart}
\small
\begin{tabular}{lcccc}
\toprule
\textbf{Method} & \textbf{Evals} & \textbf{Surr.} & \textbf{Actual} $\uparrow$ & $\boldsymbol{\Delta}$ $\downarrow$ \\
\midrule
WS Raw & $540$ & $0.874$ & $\underline{0.808}$ & $\underline{0.065}$ \\
WS Flow & $540$ & $0.867$ & $0.773$ & $0.095$ \\
WS VAE & $540$ & $0.862$ & $\mathbf{0.828}$ & $\mathbf{0.034}$ \\
\midrule
CMA-ES Raw & $540$ & $0.867$ & $0.788$ & $0.079$ \\
CMA-ES Raw & $5{,}000$ & $0.922$ & $0.781$ & $0.141$ \\
\bottomrule
\end{tabular}
\end{table}

\subsection{Ablation: Surrogate Configuration}
\label{sec:ablation}

Our default pipeline evaluates all methods through the MLP surrogate ($R^2{=}0.937$) by decoding latent codes back to parameters. An alternative, which is common practice in latent space optimization~\cite{gomez2018automatic, maus2022lolbo, lee2025nfbo}, is to let each latent model use its own prediction head directly, avoiding the decode step. Table~\ref{tab:ablation} shows this degrades performance substantially: GC-PPO Flow drops from $0.642$ to $0.478$ ($-26\%$), and GC-PPO VAE from $0.485$ to $0.458$. The prediction heads have lower accuracy ($R^2{=}0.913$ for the flow head, $0.854$ for the VAE head) because they are trained from compressed $4$D inputs with a multi-task loss rather than from the full $9$D input with a dedicated objective. The gap also increases (flow: $0.012 \to 0.050$), indicating that the lower-accuracy predictor enables more surrogate exploitation. This finding has a practical implication: when latent-space methods are compared against methods using a separate, higher-accuracy evaluator, observed performance differences may reflect evaluator quality rather than representation quality. The shared surrogate approach avoids this by ensuring all methods operate against the same evaluator.

\begin{table}[t]
\centering
\caption{Ablation: surrogate configuration ($10$ held-out targets). Shared: all methods use the MLP surrogate. Own head: each model uses its prediction head.}
\label{tab:ablation}
\small
\begin{tabular}{lccc}
\toprule
\textbf{Method} & \textbf{Surrogate} & \textbf{Actual} $\uparrow$ & $\boldsymbol{\Delta}$ $\downarrow$ \\
\midrule
GC-PPO Flow & Shared MLP & $\mathbf{0.642}$ & $\mathbf{0.012}$ \\
GC-PPO Flow & Own head & $0.478$ & $0.050$ \\
\midrule
GC-PPO VAE & Shared MLP & $\underline{0.485}$ & $0.025$ \\
GC-PPO VAE & Own head & $0.458$ & $\underline{0.023}$ \\
\bottomrule
\end{tabular}
\end{table}
\section{Discussion}
\label{sec:discussion}

\paragraph{The role of latent spaces in generalized inversion.}
The flow latent space provides substantial benefit for goal-conditioned inversion ($0.642$ vs.\ $0.523$ in the original space, a $23\%$ improvement) when all methods use the same surrogate. The $4$D manifold reduces the action space and constrains the policy to parameter configurations that the latent model has learned to be physically meaningful, paralleling findings in humanoid motor control~\cite{luo2024pulse} where latent spaces aid generalization across diverse objectives. The flow's bijective mapping also appears to provide implicit regularization: its gap ($\Delta{=}0.012$) is the smallest of any method, suggesting the policy stays in regions where the surrogate is reliable. The VAE latent space is less effective for direct GC-PPO ($0.485$) but produces the best warm-start initialization (WS VAE $0.828$ vs.\ WS Flow $0.773$). One possible explanation is that the VAE's stochastic encoding generates more diverse initial points that seed CMA-ES in different basins, while the flow's deterministic mapping concentrates initializations more narrowly.

\paragraph{Evaluation design and surrogate exploitation.}
Our ablation (Table~\ref{tab:ablation}) shows that using each latent model's own prediction head, rather than the shared MLP surrogate, degrades GC-PPO Flow by $26\%$. This is a practical finding for the broader latent space optimization community: when comparing methods across different representation spaces, the evaluator must be controlled, or performance differences may reflect evaluator accuracy rather than representation quality. More broadly, the surrogate-to-simulator gap $\Delta$ increases monotonically with optimization intensity across all experiments (GC-PPO $\Delta < 0.05$; CMA-ES $540$ $\Delta{=}0.079$; CMA-ES $5{,}000$ $\Delta{=}0.141$). That CMA-ES with $5{,}000$ evaluations actually achieves \emph{lower} recovery than CMA-ES with $540$ evaluations ($0.781$ vs.\ $0.788$) underscores that more search is not always better when optimizing against an imperfect surrogate.

\paragraph{Toward food material identification.}
This work provides the inverse estimation component of a broader pipeline for food material identification. In a complete system, visual tracking methods~\cite{wu2025fruitninja, xie2024physgaussian} would extract behavioral features from video of food manipulation, and the goal-conditioned policy would instantly estimate the corresponding material parameters. The policy's speed (${\sim}10$\,ms per estimate) makes it suitable for real-time or interactive applications. The ${\sim}9$-minute training cost is a one-time investment that generalizes across all future targets.

\paragraph{Limitations.}
We evaluate on a single test case (orange peeling with AnisoMPM); generalization to other food materials and fracture modes is untested. The validation uses $10$ held-out targets with high per-target variance (std $0.084$--$0.162$), limiting statistical power for fine-grained comparisons between methods with similar means. The $9$-dimensional parameter space is modest compared to spatially varying material fields encountered in practice. The warm-start advantage over equal-budget CMA-ES is inconsistent across representations (WS VAE outperforms but WS Flow underperforms CMA-ES $540$), and should not be treated as a universal improvement.

\paragraph{Future work.}
Replacing hand-designed behavioral features with learned representations from video would close the loop between visual observation and inverse estimation. Extending to higher-dimensional parameter spaces would test whether the latent space advantage grows with dimensionality. Incorporating predictive uncertainty into the surrogate could provide a mechanism for detecting and mitigating surrogate exploitation.

\section{Conclusion}
\label{sec:conclusion}

We presented a goal-conditioned RL approach to inverse material parameter estimation for non-differentiable food fracture simulation. After establishing that RL matches evolutionary search on fixed objectives, we trained a goal-conditioned PPO policy that generalizes across arbitrary behavioral targets, producing inverse estimates in $8$ evaluations (${\sim}10$\,ms) with no per-target retraining. Operating in a normalizing flow latent space with a shared surrogate evaluator, the policy achieves $0.642$ actual recovery, outperforming the original $9$D parameter space by $23\%$. A warm-start hybrid that refines the policy's output with brief CMA-ES search achieves the best overall recovery ($0.828$, $540$ evaluations). An ablation reveals that the common practice of using model-specific prediction heads degrades performance due to surrogate exploitation, underscoring the importance of evaluation design. These findings provide a practical framework for inverse food physics and a foundation for vision-driven material identification from video observations.

{
    \small
    \bibliographystyle{ieeenat_fullname}
    \bibliography{main}
}


\end{document}